\documentclass[11pt,a4paper]{article}
\usepackage[hyperref]{emnlp2020}
\usepackage{times}
\usepackage{latexsym}
\usepackage{graphicx}

\usepackage{stackengine,scalerel}

\usepackage{microtype}
\usepackage{mathtools}
\aclfinalcopy 

\title{EmoGraph: Capturing Emotion Correlations using Graph Networks}

\author{Peng Xu\thanks{$^*$ Equal contributions.}~, Zihan Liu$^*$, Genta Indra Winata, Zhaojiang Lin, Pascale Fung \\
Center for Artificial Intelligence Research (CAiRE)\\
Department of Electronic and Computer Engineering\\
The Hong Kong University of Science and Technology, Clear Water Bay, Hong Kong\\
\texttt\{pxuab, zliucr, giwinata, zlinao\}@connect.ust.hk}

\date{}

\begin{document}
\maketitle
\begin{abstract}
Most emotion recognition methods tackle the emotion understanding task by considering individual emotion independently while ignoring their fuzziness nature and the interconnections among them. In this paper, we explore how emotion correlations can be captured and help different classification tasks. We propose EmoGraph that captures the dependencies among different emotions through graph networks. These graphs are constructed by leveraging the co-occurrence statistics among different emotion categories. Empirical results on two multi-label classification datasets demonstrate that EmoGraph outperforms strong baselines, especially for macro-F1. An additional experiment illustrates the captured emotion correlations can also benefit a single-label classification task.
\end{abstract}

\section{Introduction}


Understanding human emotions is considered as the key to building engaging dialogue systems ~\cite{zhou2018design}. However, most works on emotion understanding tasks treat individual emotions independently while ignoring the fuzziness nature and the interconnections among them. A psychoevolutionary theory proposed by \citet{plutchik1984emotions} shows that different emotions are actually correlated, and all emotions follow a circular structure. For example, ``optimism'' is close to ``joy'' and ``anticipation'' instead of ``disgust'' and ``sadness''. Without considering the fundamental inter-correlation between them, the understanding of emotions can be unilateral, leading to sub-optimal performance. These understanding can be particularly important for low resource emotions, such as ``surprise" and ``trust" whose training samples are hard to get. 
Therefore, the research question we ask is, \textit{how can we \textbf{obtain} and \textbf{incorporate} the emotion correlation to improve emotion understanding tasks, such as classification?}

To obtain emotion correlations, a possible way is to take advantage of a multi-label emotion dataset. Intuitively, emotions with high correlations will be labeled together, and therefore, emotion correlations can be extracted from the label co-occurrences. Recently, a multi-label emotion classification competition \cite{mohammad2018semeval} with 11 emotions has been introduced to promote research into emotional understanding. To tackle this challenge, the best team ~\cite{baziotis2018ntua} first pre-trains on a large amount of external emotion-related datasets and then performs transfer learning on this multi-label task. However, they still neglect the correlations between different emotions.
In this paper, we propose \textbf{EmoGraph} which leverages graph neural networks to model the dependencies between different emotions. 
We take each emotion as a node and first construct an emotion graph based on the co-occurrence statistics between every two emotion classes. Graph neural networks are then applied to extract the features from the neighbours of each emotion node.
We conduct experiments on two multi-label emotion classification datasets. Empirical results show that our model outperforms strong baselines, especially for macro-F1 score. The analysis shows that low resource emotions, such as ``trust", can particularly benefit from the emotion correlations. An additional experiment illustrates that the captured emotion correlations can also help the single-label emotion classification task. 

\section{Related Work}
For emotion classifications, \citet{Tang2016} proposed sentiment embeddings that incorporate sentiment information into word vectors. \citet{felbo2017using} trained a huge LSTM-based emotion representation by predicting emojis. Various methods have also been developed for automatic constructions of sentiment lexicons using both a supervised and unsupervised method \cite{wang2017sentiment}. \citet{duppada2018seernet} combined both pretrained representations and emotion lexicon features, which significantly improved the emotion understanding systems.
\citet{park2018plusemo2vec,fung2018empathetic,xu2018emo2vec} encoded emotion information into word representations and demonstrated improvements over emotion classification tasks. \citet{shin2019happybot,lin2019moel,xu2019clickbait} further introduced emotion supervision into generation tasks.

Multi-label classification is an important yet challenging task in natural language processing. Binary relevance \cite{boutell2004learning} transformed the multi-label problem into several independent classifiers. Other methods to model the dependencies have since been proposed by creating new labels \cite{tsoumakas2007multi}, using classifier chains \cite{read2011classifier}, graphs \cite{li2015sentence}, and RNN \cite{chen2017ensemble,yang2018sgm}. These models are either non-scalable or modeling the labels as a sequence. 

Graph networks have been applied to model relations across different tasks such as image recognition \cite{chen2019multi,garcia2017few}, and text classification \cite{ghosal2019dialoguegcn,yao2019graph} with different graph networks \cite{kipf2016semi,velivckovic2017graph}.

Despite the growing interests in low-resource studies in machine translation~\cite{artetxe2017unsupervised,lample2017unsupervised}, dialogue systems~\cite{bapna2017towards,liu2019zero,liu2020coach}, speech recognition~\cite{miao2013deep,thomas2013deep,winata2020learning}, emotion recognition~\cite{haider2020emotion}, and etc, emotion detection for low-resource emotions has been less studied.

\section{Methodology}
In this section, we first introduce the emotion graphs and then our emotion classification models. We denote the input sentence as $x$ and the emotion classes as $e = \{e_1, e_2, \cdots, e_n\}$. The label for $x$ is $y$, where $y \in \{0,1\}^n$ and $y_j$ denotes the label for $e_j$. The embedding matrix is $E$. The co-occurrence matrix of these emotions is $M$. 

\subsection{Emotion Graphs}
We take each emotion class as a node in the emotion graph. To create the connections between emotion nodes, we use the co-occurrence statistics between emotions. The intuition is that if two emotions co-occurs frequently, they will have a high correlation. Directly using this co-occurrence matrix as our graph may be problematic because the co-occurrence matrix is symmetric while the emotion relation is not symmetric. For example, in our corpus, ``anticipation" co-occurs with ``optimism" 197 times, while ``anticipation" appears 425 times and ``optimism" appears 1143 times. Thus, knowing ``anticipation" and ``optimism" co-occur is notably more important for ``anticipation" than ``optimism". Thus, we calculate the co-occurrence matrix $M$ from a given emotion corpus and then normalize $M_{i,j}$ with $M_{i,i}$ so that the graph encodes the asymmetric relation between different emotions.
\begin{equation}
    G_{i, j}^1 = \frac{M_{i, j}}{M_{i,i}}.
\end{equation}

Due to the fuzziness nature of emotions, the graph matrix $G^1$ may contain some noise. Thus, we adopted the approach in \citet{chen2019multi} to binarize $G^1$ with a threshold $\mu$ to reduce noise and tune another hyper-parameter $w$ to mitigate the over-smoothing problem \cite{li2018deeper}:
\begin{align}
    G_{i, j}^2 &= 
\begin{dcases}
    1,     & \text{if } G_{i,j}^1 \geq \mu \\
    0,      & \text{otherwise}. 
\end{dcases} \\
    G_{i, j} &= 
\begin{dcases}
    G_{i,j}^2 / (\sum_{j=1}^{n} G_{i,j}^2) ,      & \text{if i} \neq \text{j}  \\
    1 - w,  & \text{otherwise}.
\end{dcases}
\end{align}

\subsection{Emotion Classification Models}
Our emotion classification model consists of an encoder and graph-based emotion classifiers following the framework of \citet{chen2019multi}. For the encoder, we choose the Transformer (TRS) ~\cite{vaswani2017attention} and the pre-trained model BERT~\cite{devlin2019bert} for their strong representation capacities on many natural language tasks. We denote the encoded representation of $x$ as $s$. For graph-based emotion classifiers, we experiment with two types of graph networks: GCN ~\cite{kipf2016semi} and GAT ~\cite{velivckovic2017graph}. The GCN takes the features of each emotion node as inputs and applies a convolution over neighboring nodes to generate the classifier for $e_i$:
\begin{equation}
    C = \text{ReLU}(\tilde{G} E_e W_1), \label{eq4}
\end{equation}
where $\tilde{G}$ is the normalized matrix of $G$ following \citet{kipf2016semi}. $E_e$ is the embeddings for all emotions $e$, and $W_1$ is the trainable parameters. Our emotion classifiers are $C = {C_1, C_2, \cdots, C_n}$, where $C_i$ is the classifier for $e_i$. Alternatively, GAT takes the features of each node as input and learns a multi-head self-attention over the emotion nodes to generate classifiers $C$.

We then simply take the inner product between $s$ and $C_i$ to compute $\hat{y_i} $, the logits of the emotion class $e_i$ for classification:
\begin{equation}
    \hat{y_i}  = s * C_i.
    \label{yi}
\end{equation}
As our task is a multi-label prediction problem, we add a sigmoid activation to  $\hat{y_i}$ and use a cross-entropy loss function.

\section{Experimental Setup}
\subsection{Dataset and Evaluation Metrics}
We choose two datasets for our multi-label classification training and evaluation. SemEval-2018 ~\cite{mohammad2018semeval} contains 10,983 tweets with 11 different emotion categories. It is divided into three splits: training set (6838 samples), validation set (886 samples), and testing set (3259 samples). 
Due to its small label quantities and small dataset size, we crawled another Twitter dataset where we use emojis as emotion labels. We follow the same 64 emoji types as in \citet{felbo2017using} and collect 4 million twitter data, where each tweet has at least two emojis. We split them into 2.8 million (70\%) for training, 0.4 million (10\%) for validation and 0.8 million (20\%) for testing. 

Following the metrics in \citet{mohammad2018semeval}, we use Jaccard accuracy, micro-average F1-score (micro-F1), and macro-average F1-score (macro-F1) as our evaluation metrics.  


\subsection{EmoGraph and Baselines}
Our EmoGraph has four variants,\textbf{TRS-GCN},\textbf{TRS-GAT},\textbf{BERT-GCN}, and \textbf{BERT-GAT}, depending on the choice of sentence encoder (TRS/BERT) and graph networks (GCN/GAT). We compare our model to several strong baselines. \textbf{NTUA-SLP}~\cite{baziotis2018ntua} is the top-1 system of the SemEval-2018 competition, which pre-trains the model on large amounts of external emotion-related datasets. \textbf{DATN} ~\cite{yu2018improving} is the system that transfers sentiment information with dual attention transfer network.
\textbf{SGM} ~\cite{yang2018sgm} models labels as a sequences and generates the labels using beam search. \textbf{TRS} is the system that adds a linear layer on top of the Transformer~\cite{vaswani2017attention}. \textbf{BERT} is the  system that adds a linear layer on top of the BERT~\cite{devlin2019bert}.

\begin{table}[t]
\centering
\resizebox{0.45\textwidth}{!}{
\begin{tabular}{cccc}
\hline
         & Accuracy & Micro-F1 & Macro-F1 \\ \hline
NTUA-SLP & \bf 58.8    & \bf 70.1    & 52.8    \\ 
DATN  & 58.3 & - & \bf 54.4 \\ \hline
SGM      & 48.2    & 57.5    & 41.1    \\ 
TRS      & 51.1    & 63.4    & 46.7    \\
TRS-GAT  & 51.7       & 64.6    & 48.3          \\
TRS-GCN  & 51.9    & 63.8    & 49.2    \\
BERT     & 58.4    & 70.1    & 53.8    \\
BERT-GAT & 58.3    & 69.9    & \bf 56.9    \\
BERT-GCN & \bf 58.9    & \bf 70.7   &  56.3 \\ \hline
\end{tabular}
}
\caption{Comparisons among different systems on SemEval-2018 dataset.}
\label{table:compare_semeval}
\end{table}

\begin{table}[t]
\centering
\resizebox{0.43\textwidth}{!}{
\begin{tabular}{cccc}
\hline
        & Accuracy & Micro-F1 & Macro-F1 \\ \hline
TRS     & 30.4   & 46.5   & 35.0   \\
SGM     & 33.5   & 42.9   & 25.2   \\
TRS-GAT & 33.1   & 49.1   & 38.6   \\
TRS-GCN & \bf 34.4   & \bf 50.5   & \bf 40.8   \\ \hline
\end{tabular}
}
\caption{Comparisons among different systems on Twitter dataset.}
\label{table:compare_twitter}
\end{table}

\begin{table*}[t]
\centering
\resizebox{0.95\textwidth}{!}{
\begin{tabular}{ccccccccc}
\hline
      F1 score   & \# Parameters & Anger & Fear & Happiness & Surprise & Sadness & Average F1 & Accuracy   \\ \hline
LSTM & 1.11M & 72.7  & 12.2  & 53.7  & 31.3 & 69.3  & 46.2    & 66.7   \\
LSTM-GCN & 1.16M & \bf 75.6  & \bf 39.1  & \bf 60.8  & \bf 38.0 & \bf 70.5 & \bf 54.7  & \bf 69.5   \\ \hline
BERT & 110M & 83.4  & 39.1  & 69.9  &  47.1  & 80.8  & 62.0   & 78.5   \\
BERT-GCN & 110M & \bf 83.8  & \bf 51.6  & \bf 72.9  & \bf 48.7  & \bf 82.0 & \bf 65.4  & \bf 79.8   \\ \hline
\end{tabular}
}
\caption{Comparisons of F1 scores and accuracy between EmoGraph w/ and w/o graph on IEMOCAP dataset. }
\label{table:compare_iemocap}
\end{table*}
\section{Results and Analysis}

\paragraph{Results on two emotion classification datasets} The results on SemEval-2018 dataset are shown in Table \ref{table:compare_semeval}, which illustrates several points. Firstly, BERT-GCN/TRS-GCN consistently improves over the baseline of BERT/TRS, in terms of accuracy (+0.5\%/+0.8\%), micro-F1 (+0.6\%/+0.4\%) and macro-F1 (+2.5\%/+2.5\%). It shows the effectiveness of the emotion graph by giving the performance boost on the strong baseline BERT, especially on the macro-F1 score.
Secondly, our BERT-GCN achieves a 58.9\% accuracy score, 70.7\% micro-F1 score, and 56.3\% macro-F1 score, which beats the best system NTUA-SLP in terms of all metrics, without using any affect features or pre-training on emotional datasets. 
Thirdly, EmoGraph is particularly effective in macro-F1. For example, BERT-GAT is 2.5\% better than DATN and 3.1\% better than BERT.

We notice that for both accuracy and micro-F1, the improvements of graph networks are very marginal on the SemEval-2018 dataset. This is because the small emotion label space (only 11 emotions) limits the effectiveness of emotion graphs.
Thus, we train our models on another Twitter dataset with 64 emoji labels. Table \ref{table:compare_twitter} shows both TRS-GCN and TRS-GAT consistently improves TRS with a large margin for all metrics, which shows that graph networks are better at dealing with rich emotion information.\footnote{Data statistics, training details, and ablation studies for threshold $\mu$ (Eq. 2) and weight $w$ (Eq. 3) are in the appendix.}


\begin{table}[t]
\centering
\resizebox{0.32\textwidth}{!}{
\begin{tabular}{ccc} \hline
& Surprise & Trust  \\ \hline
BERT    & 18.7   & 6.9   \\
BERT-GAT   & \bf 31.9   &  14.8    \\ 
BERT-GCN    & 27.8      & \bf 24.3  \\ \hline
\end{tabular}
}
\caption{ Comparison of F1 scores between BERT-GAT, BERT-GCN and BERT for ``surprise" and ``trust".}
\label{table:compare_semeval_bert}
\end{table}


\begin{figure}[!ht]
    \centering
    \includegraphics[scale=0.45]{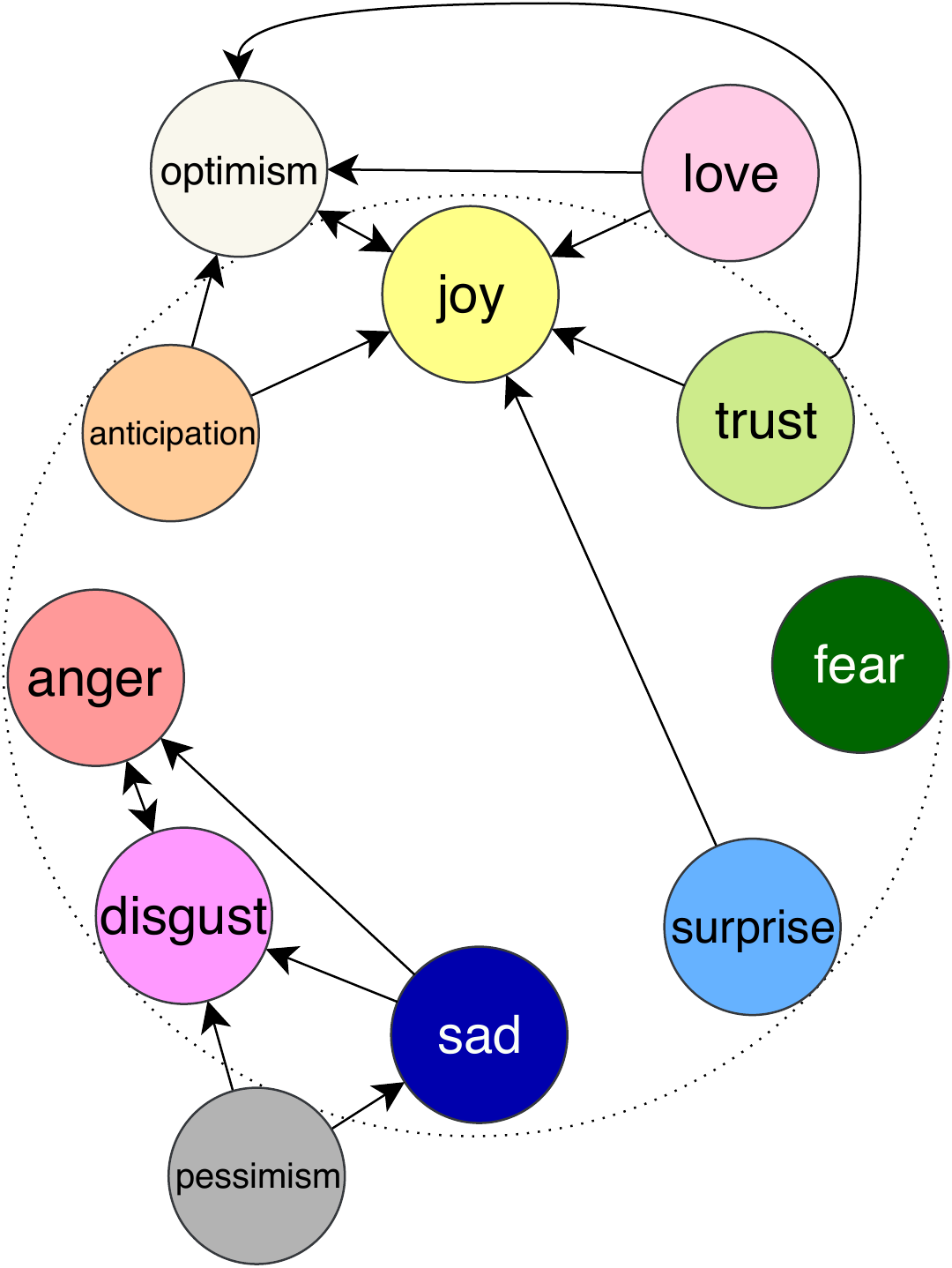}
    \caption{The graph visualization of $G$. The arrow denotes the conditional dependency from the source node to the target node.}
    \label{fig:graph-wheel}
\end{figure}

\paragraph{Connection to the wheel of emotions}
Following the positioning of the wheel of emotion~\cite{plutchik1984emotions}, we visualize the graph structure $G$ in Figure \ref{fig:graph-wheel}. It shows that our emotion graph can be divided into three sub-graphs, 1) the nodes connected with ``disgust", 2) ``fear" 3) the nodes connected with ``joy". Most nodes are locally connected, which means the positioning in the wheel of emotion does contain the correlation information. One exception is that ``anticipation" is also close to ``anger" in the wheel of emotions, which is not true in our case. Another exception we found is that ``surprise" is close to ``joy" in our emotion graph while they are quite far away from each other in the wheel of emotions. We believe it reflects the bias when people post tweets online, and the distance between neighboring emotions are not quantitatively the same as the wheel of emotions. 

\paragraph{Analysis on low resource emotions} Table \ref{table:compare_semeval_bert} shows that the significant improvements in terms of the macro-F1 on the SemEval-2018 dataset mainly come from the low resource emotions, such as ``surprise" and ``trust", where only 5\% of the labels are positive. From Figure \ref{fig:graph-wheel}, we observe that both ``surprise" and ``trust" are connected to ``joy", which has 39.3\% samples labeled as positive. We conjecture that low resource emotion classifiers learn effective inductive bias through connections with high-resource emotion classes, such as ``joy" and ``optimism", and achieve better performances. 

\paragraph{Generalization to single emotion classification task}
To verify the generalization ability of EmoGraph, we conduct an extra experiment on IEMOCAP dataset \cite{busso2008iemocap}, which is a single label multi-class classification dataset. We only use the six emotions that overlap with the SemEval-2018 task, which are ``anger", ``sadness", ``happiness", ``disgust", ``fear" and ``surprise" with 2931 samples (the graph matrix G is obtained from the SemEval-2018 dataset). We then train our EmoGraph using either a one-layer LSTM with attention or BERT as an encoder and GCN as the graph structure. To adapt the change from 11 emotions to 6 emotions, we only train the classifiers $C_i$ corresponding to those six emotions. 
The results with 10-fold cross-validation are reported in Table \ref{table:compare_iemocap}. We didn't report ``disgust" as there are less than five positive examples. The results confirm that with captured emotion correlations, classification performances can be improved by 2.8\%/8.5\% accuracy/average F1 score using LSTM encoder and 1.3\%/3.4\% accuracy/average F1 score using BERT encoder. 
Surprisingly, for both encoders with GCN improves more than 10\% on ``fear" (a low-resource emotion in the IEMOCAP dataset), although it doesn't have connections with other emotions in the graph. We conjecture that the improvements come from the shared representation of $W_1$ in Eq.~\ref{eq4} among all emotion categories, which helps our model optimize to a better local minimal.











\section{Conclusion}
In this paper, we proposed \textbf{EmoGraph} which leverages graph neural networks to model the dependencies among different emotions. We consider each emotion as a node and construct an emotion graph based on the co-occurrence statistics.
Our model with EmoGraph outperforms the existing strong multi-label classification baselines. Our analysis shows EmoGraph is especially helpful for low resource emotions and large emotion space. An additional experiment shows that it can also help a single-label emotion classification task.

\section*{Acknowledgments}
This work is partially funded by ITF/319/16FP, HKUST 16248016 of Hong
Kong Research Grants Council and MRP/055/18 of the Innovation Technology Commission, the Hong Kong SAR Government.

\bibliographystyle{acl_natbib}
\bibliography{emnlp2020}

\clearpage

\appendix

\section{Appendices}
\label{sec:appendix}
\subsection{Preprocessing}

To cope with the noise in the twitter data, we lowercase the tweets, remove the ``\#'' punctuation and replace  ``URL link" with a special token \texttt{<url>}, ``user mentions" with \texttt{<user>} and ``number" with \texttt{<num>}. For example, one original tweet \textit{"@JessicaZ00 @ZRlondon ditto!! Such an amazing atmosphere! We have 10 people here. \#LondonEvents \#cheer"} will be cleaned as \textit{"\texttt{<user>} \texttt{<user>} ditto! such an amazing atmosphere! we have \texttt{<num>} people here. londonevents cheer"}. By doing so, we can reduce the vocabulary size and make the model easier to learn.

\label{sec:supplemental}
\subsection{Training Details}
To train the SemEval-2018, we calculate the co-occurrence matrix $G^1$ based on the training and development sets. We set the threshold $\mu$ as 0.4, and the weight $w$ as 0.35 for GCN. We set the threshold $\mu$ as 0.5 for GAT. For both BERT-GCN and BERT-GAT, the emotion label embeddings are initialized with 300-dim GloVe vectors. For the BERT structure, we choose 12 layers BERT-base model, and the hidden size of the graph is set to 768. An Adam optimizer with the learning rate 1e-4 is used to train the graph networks. Another Adam optimizer is used to train the BERT model with the learning set as 2e-5 and dropout rate as 0.3. For TRS-GCN and TRS-GAT, the hidden size of the graph network is set to 200. The heads of the Transformer are 4, and the depth is 44. An Adam optimizer is used to train the full model with a 0.001 learning rate.  For the Twitter dataset, we calculate the co-occurrence matrix $G^1$ based on the training set. We set the threshold $\mu$ as 0.1, and the weight $w$ is set as 0.35 for both. We use the TRS as the sentence encoder. The hidden size of both graph networks and TRS are 200. The heads of TRS are 6, and the depth is 120. An Adam optimizer is used to train our model with a 0.001 learning rate.

To train IEMOCAP dataset, the parameters of both GCN and BERT follow the same setting as in SemEval 2018 task. For the LSTM based method, we set the hidden size of LSTM encoder as 200 and initialize the word embedding with GloVe. An Adam optimizer with a learning rate of 0.001 is used to optimize all parameters.

\begin{table}[t]
\centering
\resizebox{\linewidth}{!}{
\begin{tabular}{cccccc}
\hline
Anger                & Anticipation         & Disgust              & Fear                 & Joy                  & Love                 \\ \hline
36.1                 & 13.9                 & 36.6                 & 16.8                 & 39.3                 & 12.3                 \\ \hline
\multicolumn{1}{l}{} & \multicolumn{1}{l}{} & \multicolumn{1}{l}{} & \multicolumn{1}{l}{} & \multicolumn{1}{l}{} & \multicolumn{1}{l}{} \\ \cline{1-5}
Optimism             & Pessimism            & Sadness              & Surprise             & Trust                &                      \\ \cline{1-5}
31.3                 & 11.6                 & 29.4                 & 5.2                  & 5.0                  &                      \\ \cline{1-5}
\end{tabular}
}
\caption{Percentage of tweets that are labeled with a given emotion in the SemEval-2018 Task 1 dataset.}
\label{table:semeval_data}
\end{table}

\begin{table}[t]
\centering
\resizebox{0.89\linewidth}{!}{
\begin{tabular}{cccccccc}
\includegraphics[height=0.75cm]{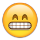} & \includegraphics[height=0.75cm]{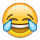} & \includegraphics[height=0.75cm]{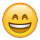} & \includegraphics[height=0.75cm]{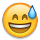} & \includegraphics[height=0.75cm]{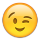} & \includegraphics[height=0.75cm]{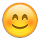} & \includegraphics[height=0.75cm]{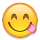} & \includegraphics[height=0.75cm]{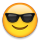} \\
33.37 & 21.86 & 38.13 & 16.63 & 0.41 & 15.40 & 3.71 & 4.35 \\ 
\includegraphics[height=0.75cm]{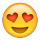} & \includegraphics[height=0.75cm]{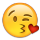} & \includegraphics[height=0.75cm]{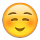} & \includegraphics[height=0.75cm]{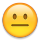} & \includegraphics[height=0.75cm]{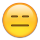} & \includegraphics[height=0.75cm]{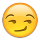} & \includegraphics[height=0.75cm]{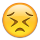} & \includegraphics[height=0.75cm]{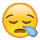} \\
32.95 & 0.95 & 3.78 & 1.43 & 5.02 & 0.21 & 3.13 & 10.13 \\
\includegraphics[height=0.75cm]{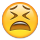} & \includegraphics[height=0.75cm]{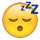} & \includegraphics[height=0.75cm]{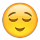} & \includegraphics[height=0.75cm]{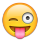} & \includegraphics[height=0.75cm]{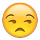} & \includegraphics[height=0.75cm]{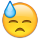} & \includegraphics[height=0.75cm]{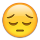} & \includegraphics[height=0.75cm]{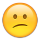}\\
2.59 & 0.26 & 9.67 & 4.40 & 2.18 & 2.83 & 4.86 & 2.23 \\
 \includegraphics[height=0.75cm]{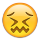} & \includegraphics[height=0.75cm]{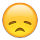} & \includegraphics[height=0.75cm]{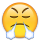} & \includegraphics[height=0.75cm]{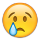} & \includegraphics[height=0.75cm]{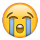} & \includegraphics[height=0.75cm]{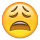} & \includegraphics[height=0.75cm]{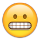} & \includegraphics[height=0.75cm]{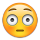} \\
1.39 & 1.58 & 3.50 & 3.03 & 1.98 & 3.35 & 1.14 & 3.73 \\
\includegraphics[height=0.75cm]{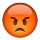} & \includegraphics[height=0.75cm]{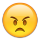} & \includegraphics[height=0.75cm]{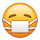} & \includegraphics[height=0.75cm]{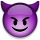} & \includegraphics[height=0.75cm]{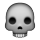} & \includegraphics[height=0.75cm]{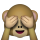} & \includegraphics[height=0.75cm]{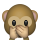} & \includegraphics[height=0.75cm]{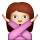} \\
2.92 & 3.21 & 9.91 & 0.51 & 2.16 & 4.00 & 3.30 & 2.97 \\
\includegraphics[height=0.75cm]{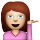} & \includegraphics[height=0.75cm]{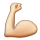} & \includegraphics[height=0.75cm]{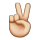} & \includegraphics[height=0.75cm]{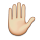} & \includegraphics[height=0.75cm]{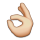} & \includegraphics[height=0.75cm]{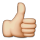} & \includegraphics[height=0.75cm]{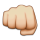} & \includegraphics[height=0.75cm]{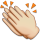} \\
5.14 & 3.66 & 1.69 & 1.89 & 2.43 & 6.95 & 2.10 & 2.56 \\
\includegraphics[height=0.75cm]{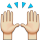} & \includegraphics[height=0.75cm]{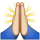} & \includegraphics[height=0.75cm]{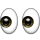} & \includegraphics[height=0.75cm]{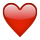} & \includegraphics[height=0.75cm]{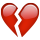} & \includegraphics[height=0.75cm]{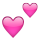} & \includegraphics[height=0.75cm]{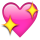} & \includegraphics[height=0.75cm]{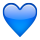} \\
0.19 & 3.83 & 1.35 & 3.49 & 1.55 & 2.12 & 2.83 & 0.78  \\
\includegraphics[height=0.75cm]{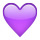} & \includegraphics[height=0.75cm]{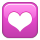} & \includegraphics[height=0.75cm]{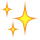} & \includegraphics[height=0.75cm]{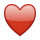} & \includegraphics[height=0.75cm]{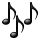} & \includegraphics[height=0.75cm]{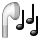} & \includegraphics[height=0.75cm]{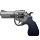} & \includegraphics[height=0.75cm]{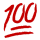} \\    
1.48 & 2.03 & 0.94 & 0.25 & 1.28 & 0.53 & 0.56 & 0.28 \\
\end{tabular}
}
\caption{Percentage of tweets that are labeled with a given emoji in the twitter data. Note that our dataset is multi-labeled, the sum of all classes is not one.}
\label{table:twitter_data}
\end{table}

\begin{table}[t!]
\centering
\resizebox{\linewidth}{!}{
\begin{tabular}{cccccc}
\hline
      F1 score   & Anger & Fear & Happiness & Surprise & Sadness \\ \hline
Distribution(\%)       & 37.7     & 1.4  & 20.1      & 3.5      & 37.2           \\ \hline
\end{tabular}}
\caption{Percentage of data samples that are labeled with a given emotion in the IEMOCAP dataset.}
\label{table:iemocap_data}
\end{table}

\subsection{Data Statistics}
The data statistics for the SemEval-2018, Twitter and IEMOCAP are shown in Table~\ref{table:semeval_data}, Table~\ref{table:twitter_data}, and Table~\ref{table:iemocap_data}, respectively.

From Table~\ref{table:semeval_data}, we can see that in the SemEval-2018 dataset, only 5.2\% and 5.0\% samples are labeled as positive for ``surprise'' and ``trust'', respectively. And from Table~\ref{table:iemocap_data}, we can see that in the IEMOCAP dataset, only 1.4\% samples are labeled as positive for ``fear''.

\begin{figure}[t]
  \centering
  \includegraphics[width=\linewidth]{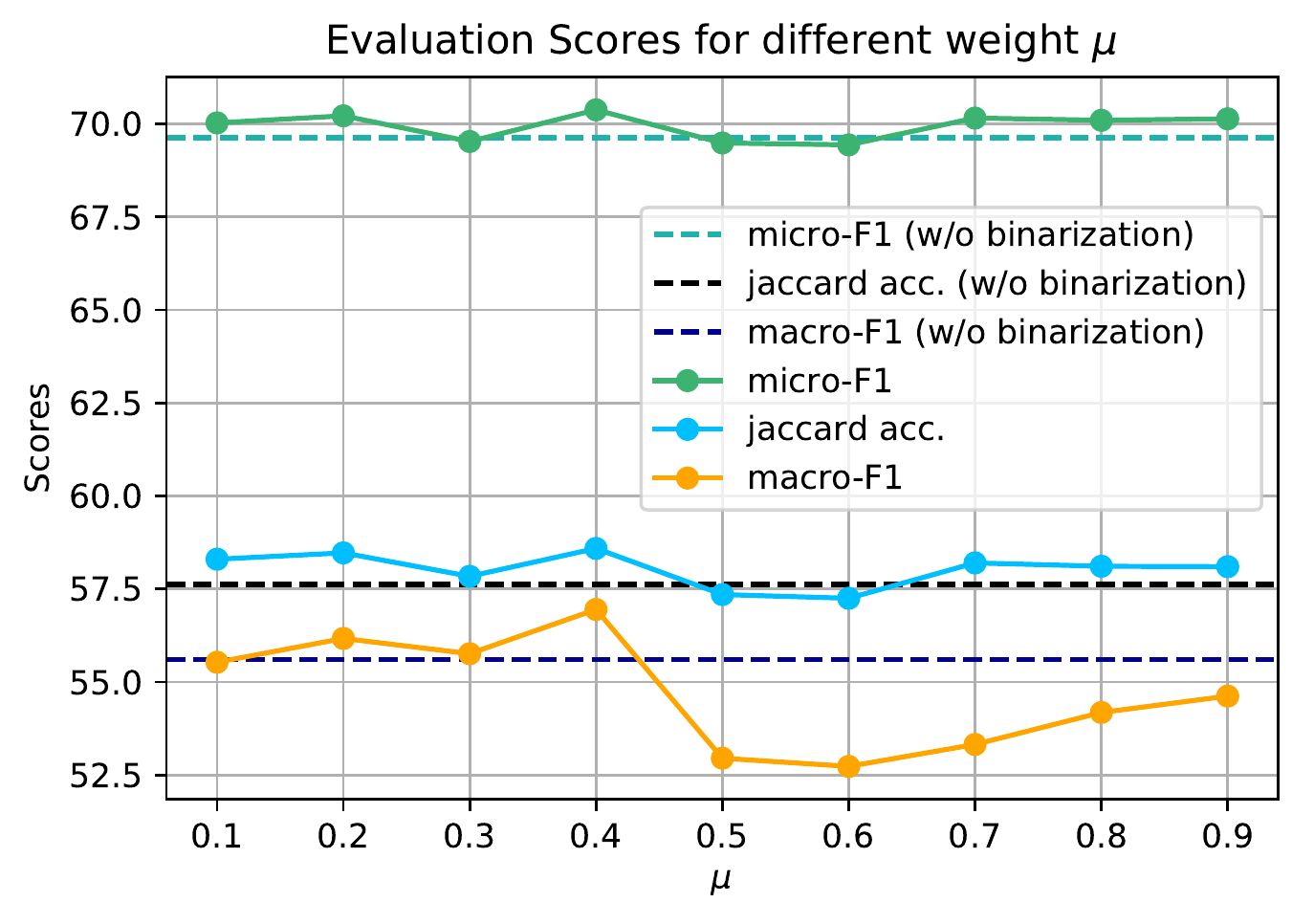}
  \caption{Evaluation Scores for different threshold $\mu$ on the SemEval-2018 dataset.}
  \label{fig:scores-mu}
  \centering
  \includegraphics[width=\linewidth]{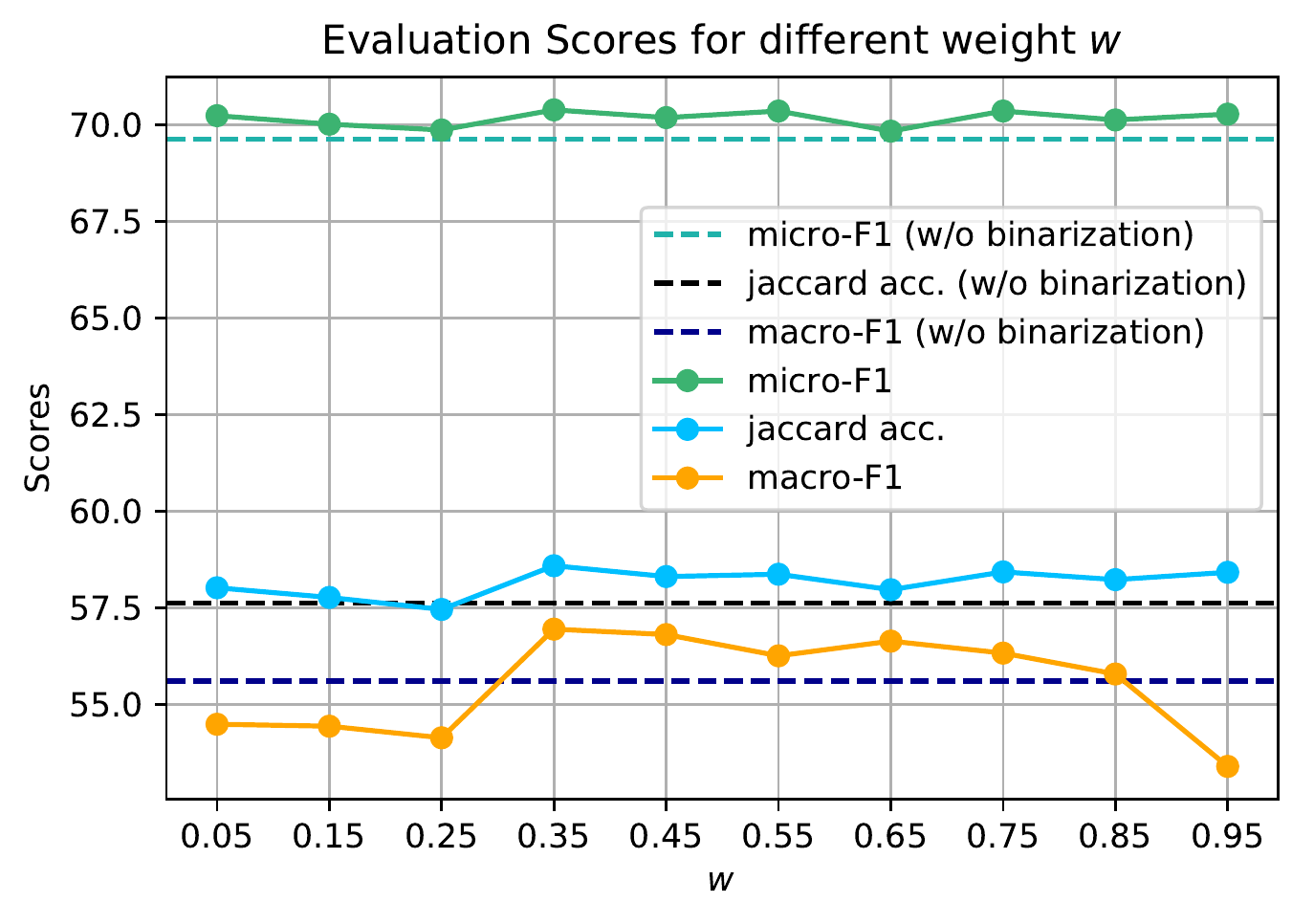}  
  \caption{Evaluation Scores for different weight $w$ on the SemEval-2018 dataset.}
  \label{fig:scores-w}
\end{figure}

\subsection{Effects of threshold $\mu$ and weight $w$}
As illustrated in Figure~\ref{fig:scores-mu} and Figure~\ref{fig:scores-w}, we plot the diagram of accuracy, micro-F1 and macro-F1 on the test set by varying $\mu$ and $w$, respectively, for BERT-GCN. We also include another baseline BERT-GCN model which is trained by setting $G = G^1$ (w/o binarization), to see the effects of binarization (Eq. 2) and weighting (Eq. 3). 

Figure~\ref{fig:scores-mu} shows that removing moderate amount of noisy connections by changing $\mu$ can improve the performance. For macro-F1, it clearly shows that small $\mu$ achieves much better results than large $\mu$. If $\mu$ is large, useful emotion connections might be cut off, therefore leading to worse results.

Figure~\ref{fig:scores-w} shows that for accuracy and micro-F1, changing $w$ can achieve better performance than the baseline without binarization and weighting.

\end{document}